\documentclass[default,iicol]{sn-jnl}% Default with double column layout

\usepackage{graphicx}%
\usepackage{multirow}%
\usepackage{amsmath,amssymb,amsfonts}%
\usepackage{amsthm}%
\usepackage{mathrsfs}%
\usepackage[title]{appendix}%
\usepackage{xcolor}%
\usepackage{textcomp}%
\usepackage{manyfoot}%
\usepackage{booktabs}%
\usepackage{algpseudocode}%
\usepackage{listings}%
\usepackage{booktabs} % For formal tables
\usepackage{xurl} % for breaks in url's
\usepackage[linesnumbered,lined,commentsnumbered,ruled,vlined]{algorithm2e}
\usepackage[ruled]{algorithm2e} % For algorithms
\usepackage[normalem]{ulem}
\usepackage{subfig}
\theoremstyle{thmstyleone}%
\theoremstyle{thmstyletwo}%

\theoremstyle{thmstylethree}%

\begin{document}

\title[Annotated Hands for Generative Models]{Annotated Hands for Generative Models}

%%=============================================================%%
%% Prefix	-> \pfx{Dr}
%% GivenName	-> \fnm{Joergen W.}
%% Particle	-> \spfx{van der} -> surname prefix
%% FamilyName	-> \sur{Ploeg}
%% Suffix	-> \sfx{IV}
%% NatureName	-> \tanm{Poet Laureate} -> Title after name
%% Degrees	-> \dgr{MSc, PhD}
%% \author*[1,2]{\pfx{Dr} \fnm{Joergen W.} \spfx{van der} \sur{Ploeg} \sfx{IV} \tanm{Poet Laureate} 
%%                 \dgr{MSc, PhD}}\email{iauthor@gmail.com}
%%=============================================================%%

% \author*[1,2]{\fnm{Yue} \sur{Yang}}\email{yyang941@gatech.edu}

% \author[2,3]{\fnm{Atith N} \sur{Gandhi}}\email{agandhi98@gatech.edu}
% \equalcont{These authors contributed equally to this work.}

% \author[1,2]{\fnm{Greg} \sur{Turk}}\email{turk@cc.gatech.edu}
% \equalcont{These authors contributed equally to this work.}

% \affil*[1]{\orgdiv{Department}, \orgname{Organization}, \orgaddress{\street{Street}, \city{City}, \postcode{100190}, \state{State}, \country{Country}}}

% \affil[2]{\orgdiv{Department}, \orgname{Organization}, \orgaddress{\street{Street}, \city{City}, \postcode{10587}, \state{State}, \country{Country}}}

% \affil[3]{\orgdiv{Department}, \orgname{Organization}, \orgaddress{\street{Street}, \city{City}, \postcode{610101}, \state{State}, \country{Country}}}

\author{\fnm{Yue} \sur{Yang}}\email{yygx@cs.unc.edu}\equalcont{These authors contributed equally to this work.}

\author{\fnm{Atith N} \sur{Gandhi}} \email{agandhi98@gatech.edu}\equalcont{These authors contributed equally to this work.}

\author*{\fnm{Greg} \sur{Turk}}\email{turk@cc.gatech.edu}

\affil{\orgdiv{College of Computing}, \orgname{Georgia Institute of Technology}, \orgaddress{\street{225 North Avenue}, \city{Atlanta}, \postcode{30332}, \state{GA}, \country{USA}}}

\abstract{Generative models such as GANs and diffusion models have demonstrated impressive image generation capabilities. Despite these successes, these systems are surprisingly poor at creating images with hands. We propose a novel training framework for generative models that substantially improves the ability of such systems to create hand images. Our approach is to augment the training images with three additional channels that provide annotations to hands in the image. These annotations provide additional structure that coax the generative model to produce higher quality hand images. We demonstrate this approach on two different generative models: a generative adversarial network and a diffusion model. We demonstrate our method both on a new synthetic dataset of hand images and also on real photographs that contain hands. We measure the improved quality of the generated hands through higher confidence in finger joint identification using an off-the-shelf hand detector.}


\maketitle

\section{Introduction}
\label{sec:intro}

Generative models have recently demonstrated impressive results, creating images of scenes that may contain a wide variety of realistic looking objects. Even the richness of human faces are captured well by generative models. One notable failing of these models, however, is the generation of hands~\cite{Uncanny, Nightmares, Struggles}. Generative models will often create hands that have too many or too few fingers, left hands in a right hand context (and vice-versa), and fingers that are bent at unnatural angles. A web search for ``AI art hands'' leads to numerous articles discussing the problem of drawing hands using generative models. Why are hand images so difficult for generative models to create? 

We speculate that there are several reasons why generative models are poor at creating hands. First, there may be relatively few images of hands in a given training dataset, or the hands may often take up only a small portion of the image. Second, hands are high degree-of-freedom objects where each of the 15 joints can be at many number of different angles. For this reason, a generative model has to learn a high dimensional manifold of possible hand shapes. Also, whereas many objects (e.g. faces, cars, houses) have a natural upright pose, hands are found in a wide variety of orientations. Finally, the non-thumb fingers of a hand look very similar to each other. This similarity of appearance could cause issues in terms of the number of fingers that are generated and for distinguishing right from left hands. 

Our approach to improving the hand generation capabilities of a generative model is to provide additional information beyond the red, green and blue color channels of an image. We create three additional channels that essentially help to ``teach'' the generative model about hands during training. These \emph{annotation channels} indicate that a hand has five fingers, that there are three separate segments on each finger, and that a hand can be left or right. We train a generative model using six-channel images, which requires the model to match not only the rgb colors of the image, but also to match the three additional annotation channels. We have found that training a generative model using such augmented images causes the generative model to produce higher quality hand images.

Our training approach is model agnostic, and can be used to train any image generation model. We demonstrate our method with StyleGAN2 and with a simple diffusion model. We show improved hand generation results in both cases, as measured by several hand quality metrics.

The key contributions in our work are as follows:

\begin{itemize}
  \item Developed a method of creating annotations of hand images for use in training generative architectures that produce improved hand images.
  \item Created a synthetic hand image \textbf{dataset} that can be used for evaluating the effectiveness of hand synthesis.
  \item Developed measures for evaluating the quality of hands in images.
  \item Demonstrated our training approach for hands can be used with both generative adversarial networks and diffusion models.
\end{itemize}

\section{Related Work}

While early autoencoder models were capable of creating simple images, the rise of generative image models really took off with the invention of generative adversarial networks (GANs) by Goodfellow et al.~\cite{goodfellow2020generative}. Soon researchers were able to create powerful GAN models with architectures such as BigGAN~\cite{brock2018large} and Progressive GAN~\cite{karras2017progressive}. Further advances in the form of the StyleGAN architecture enabled more structured latent spaces to be trained~\cite{karras2019style,karras2020analyzing}, which gave users more control over image generation.

Variational autoencoders provided a new direction for generative models~\cite{kingma2013auto}. In particular, vector quantized variational autoencoders (VQ VAE) proved to be a particularly powerful tool for image generation~\cite{van2017neural,esser2021taming}. Through contrastive learning, the CLIP model learned a shared latent space for both text and images~\cite{radford2021learning}. By combining CLIP and a VQ VAE, the original DALL-E architecture allowed a user to create images based on text prompts~\cite{ramesh2021zero}.

Recently, diffusion models have proved to be highly effective at image generation~\cite{ho2020denoising,dhariwal2021diffusion}.  Classifier-free guidance of diffusion models proved to be an effective way of boosting the quality of images created by diffusion models~\cite{ho2022classifier}. DALL-E 2 demonstrated the control of a diffusion model using text prompts using CLIP embeddings, and these results drew considerable attention to diffusion models~\cite{ramesh2022hierarchical}. DALL-E 3 improved the generated image quality by training with improved image captions~\cite{shi2020improving}. Rombach et al. introduced a latent space encoding of pixels that allows high resolution images to be generated by diffusion at a lower computational cost~\cite{rombach2022high}, and this approach was used to train the popular Stable Diffusion model.

A variety of techniques have been introduced to control diffusion models. Textual inversion and Dreambooth are two approaches that allows a diffusion model to be tuned to generate images of a particular object~\cite{gal2022image,ruiz2023dreambooth}. Prompt-to-prompt is a technique that leverages control of the cross-attention layers to allow a user to tune an image by performing edits to their text prompt~\cite{hertz2022prompt}. Instruct Pix-2-Pix trains a diffusion model based on an image editing suite, and the resulting model allows text edits to images~\cite{brooks2023instructpix2pix}. ControlNet fine-tunes a diffusion model to allow control over image generation using images such as one that contain edges, depth, or color-coded body parts~\cite{zhang2023adding}.

\section{Preliminaries}

\subsection{Generative Adversarial Networks}

Generative Adversarial Networks (GANs), introduced by~\citeauthor{goodfellow2014generative}, were a significant advance in generative models. GANs consist of two neural networks, a generator (G) and a discriminator (D), engaged in a minimax game. The generator strives to create data that is indistinguishable from real samples, while the discriminator aims to correctly classify whether a given sample is real or generated. This adversarial training process is driven by the following loss function:

\begin{equation}
\begin{split}
\min_G \max_D V(D, G) = \mathbb{E}_{\mathbf{x} \sim p_{\text{data}}(\mathbf{x})} [\log D(\mathbf{x})] +\\ \mathbb{E}_{\mathbf{z} \sim p_{\mathbf{z}}(\mathbf{z})} [\log(1 - D(G(\mathbf{z})))]
\end{split}
\end{equation}

Here, $p_{\text{data}}(\mathbf{x})$ represents the distribution of real data, and $p_{\mathbf{z}}(\mathbf{z})$ is the distribution of noise variables. GANs excel at generating high-quality, novel data and have applications across various domains, from image synthesis to natural language processing.

In this work, we are using StyleGAN2 \cite{karras2019style} which is an update over the first StyleGAN \cite{karras2020analyzing}. StyleGAN is a generative adversarial network whose design is inspired by style transfer \cite{huang2017arbitrary}. The major contribution of StyleGAN paper was to create a less entangled latent space. This along with other modifications in the generative adversarial network resulted in the better quality image generation. Further modifications in the generator were made in the StyleGAN2 to avoid generation of blob-like artifacts in the image, and improve the overall quality. 

\subsection{Diffusion Models}
% Diffusion models have garnered significant attention recently, where a neural network learns to progressively denoise data, starting from pure noise, and finally generate realistic images. The setup of a denoising diffusion model involves two crucial processes:
% \begin{enumerate}
%     \item A fixed (or predefined) forward diffusion process, denoted as \(q\), is chosen. This process gradually introduces Gaussian noise to an image, starting from an initial state of pure noise.
%     \item A learned reverse denoising diffusion process, denoted as \(p_\theta\), is employed. In this process, a neural network is trained to gradually denoise an image that begins as pure noise and progresses until it converges to an actual image.
% \end{enumerate}

% Both the forward and reverse processes, indexed by \(t\), occur over a finite number of time steps, \(T\), typically a large value (e.g., \(T = 1000\)). The process initiates at \(t = 0\), where a real image \(\mathbf{x}_0\) is sampled from the data distribution (e.g., an ImageNet image of a cat). The forward process introduces Gaussian noise at each time step, which is added to the image from the previous step. With a well-structured schedule for adding noise, this gradual process eventually leads to the convergence to what is known as an isotropic Gaussian distribution at \(t = T\).

Recently, diffusion models have garnered significant attention due to their ability to generate high quality images based on text prompts. These models start with pure noise and gradually generate realistic images~\cite{ho2020denoising}. The denoising diffusion model setup involves two essential processes:

\begin{enumerate}
    \item \textbf{Forward Diffusion Process:} A fixed (or predefined) forward diffusion process, denoted as \(q\), is selected. This process introduces Gaussian noise to an image, starting from an initial image from the training set.
    \item \textbf{Reverse Denoising Diffusion Process:} A learned reverse denoising diffusion process, denoted as \(p_\theta\), is employed. In this process, a neural network is trained to progressively denoise an image that begins as pure noise and evolves until it converges to a real image.
\end{enumerate}

Both the forward and reverse processes, indexed by \(t\), unfold over a finite number of time steps, typically a large value such as \(T = 1000\). The process begins at \(t = 0\) by sampling a real image \(\mathbf{x}_0\) from the data distribution (e.g., an ImageNet cat image). At each time step, the forward process introduces Gaussian noise, which is added to the image from the previous step. This gradual noise addition, following a well-structured schedule, eventually leads to convergence to what is known as an isotropic Gaussian distribution at \(t = T\).

During training, the neural network (typically a U-Net) is trained to perform the reverse process, namely to take a partially noised image and take a small step towards a less noisy image. After the network is trained, image generation is carried out by sampling from the Gaussian distribution, and then performing many (small) steps of denoising until a fully denoised image has been created. While early diffusion methods required many steps to create an image sample, the denoising diffusion implicit model (DDIM) can generate high quality images using a small number of steps~\cite{song2020denoising}. By conditioning the denoising process based on text, a diffusion model can be created that generates images based on a text prompt~\cite{ramesh2022hierarchical}.

\section{Methods}
\label{sec:methods}
% Overview
% 1. intro the 3 additional annotation channel (Point out that how to obtain keypoints & left/right info is important)
% 2. intro the way we get keypoints & left/right info for synthesized hand
% 3. intro the way we get keypoints & left/right info for real hand
% 4. intro the way we design metrics to evaluate the quality of generated images

In this section, we provide a comprehensive introduction to our method. First, we present the methodology for creating the additional annotation channels, which is the core of our approach (Section~\ref{subsec:annotation_channels}). Next, we describe the process of creating annotation channels on two different types of data: synthesized hands (Section~\ref{subsec:synthesized_hands}) and real hands (Section~\ref{subsec:real_hands}).

\begin{figure*}[t]
\centering
\begin{minipage}[t]{.29\textwidth}
  \centering
  \includegraphics[width=0.8\columnwidth]{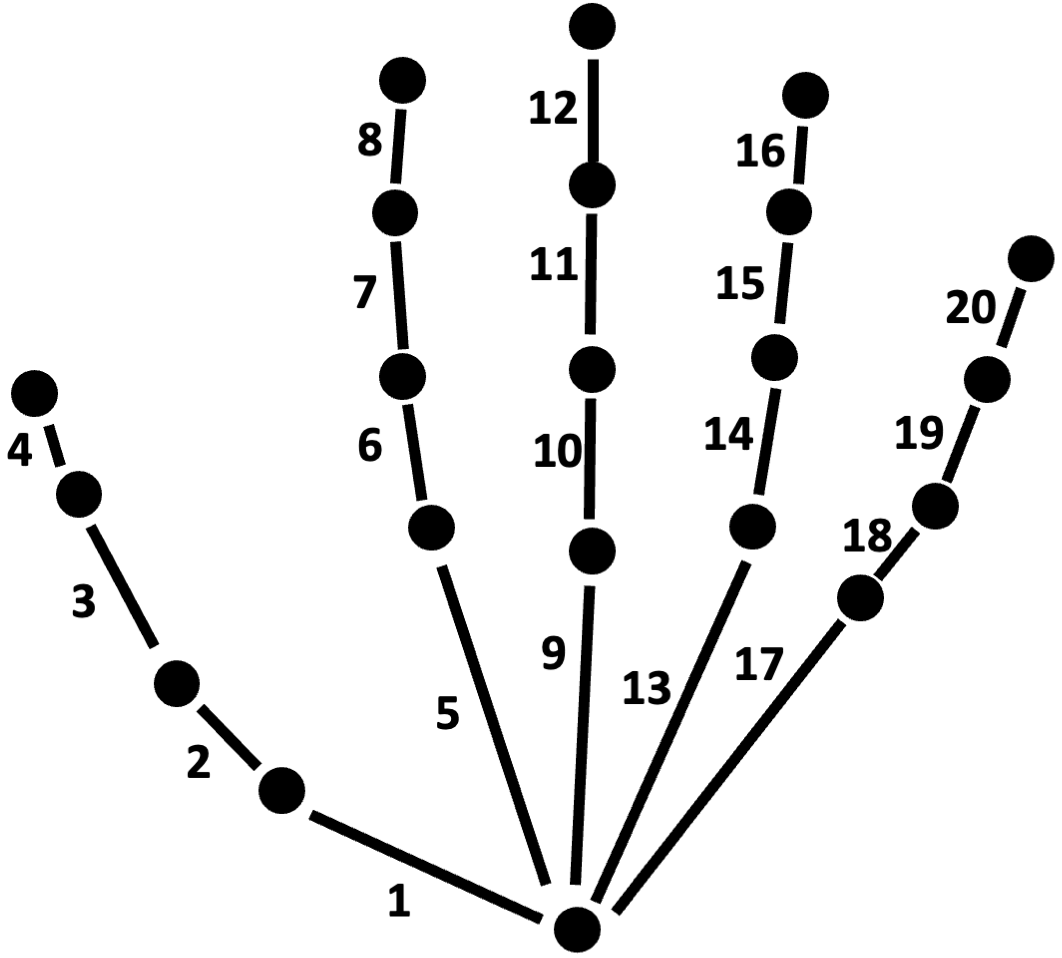}
  \caption{Hand Skeleton.}
  \label{fig:hand_skeleton}
\end{minipage}
\hspace{.1cm}
\begin{minipage}[t]{.29\textwidth}
  \centering
  \includegraphics[width=0.8\columnwidth]{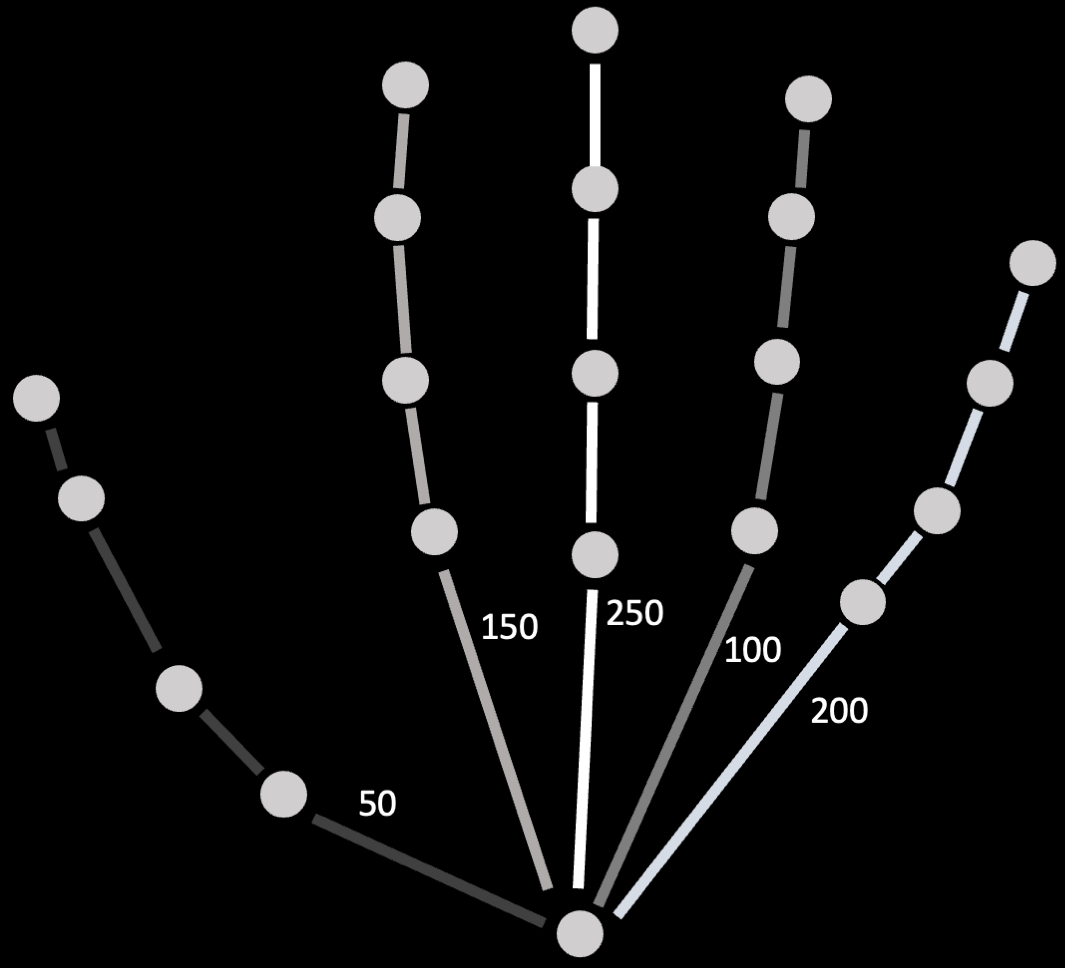}
  \caption{First Annotation Channel. }
  \label{fig:1_channel}
\end{minipage}
\hspace{.1cm}
\begin{minipage}[t]{.29\textwidth}
  \centering
  \includegraphics[width=0.8\columnwidth]{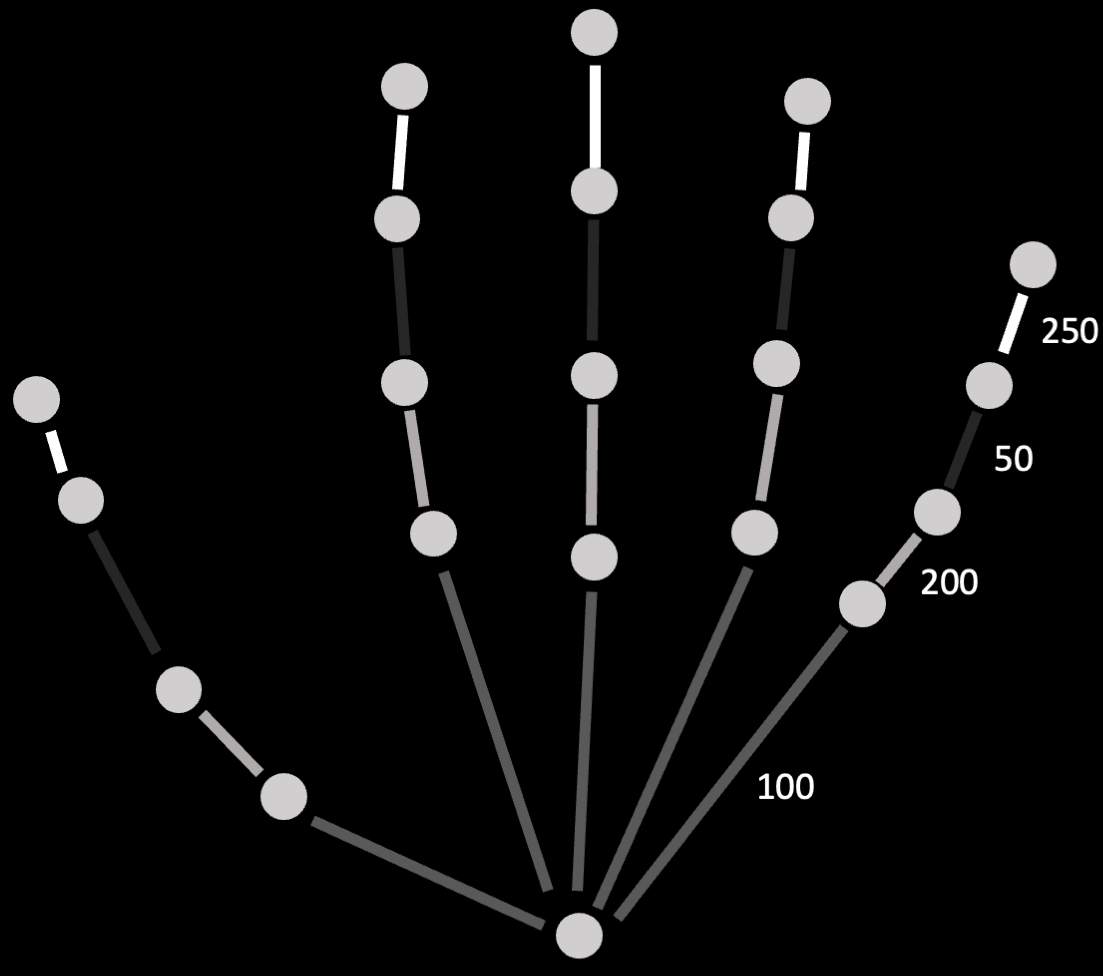}
  \caption{Second Annotation Channel.}
  \label{fig:2_channel}
\end{minipage}
\hspace{.1cm}
% \begin{minipage}[t]{.23\textwidth}
%   \centering
%   \includegraphics[width=0.8\columnwidth]{dor_ven_comp_crop.png}
%   \caption{Left and right hands share the same skeleton but different hand details.}
%   \label{fig:dor_ven_comp}
% \end{minipage}
% \vspace{-15pt}
\end{figure*}

\begin{figure*}
  \centering
  \subfloat{
    \centering
    \includegraphics[width=0.49\textwidth]{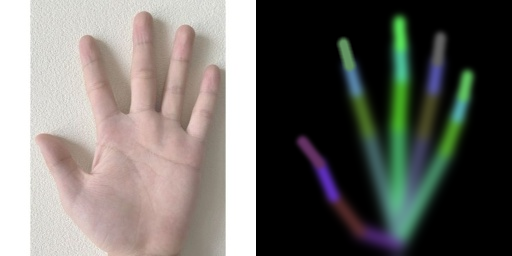}
    \label{fig:image_a}
  }
  \subfloat{
    \centering
    \includegraphics[width=0.49\textwidth]{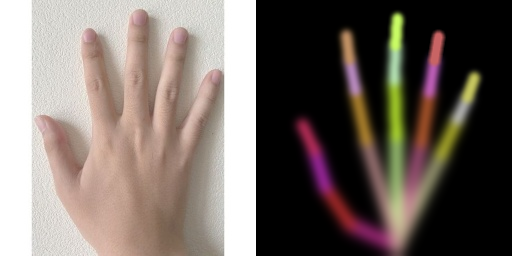}
    \label{fig:image_b}
  }
  \caption{(a) Left hand and its annotation channels. (b) Right hand and its annotation channels. }
  \label{fig:hand_annotation}
\end{figure*}

% \begin{figure*}[!tbp]
%   \centering
%   \subfloat[]{\includegraphics[width=0.2\textwidth]{merged_image_left.png}\label{fig:f1}}
%   \hfill
%   \subfloat[Flower two]{\includegraphics[width=0.2\textwidth]{merged_image_right.png}\label{fig:f2}}
%   \caption{My flowers.}
% \end{figure*}
% \begin{figure*}[t]
% \subcaptionbox*{First subfigure}[.45\linewidth]{%
%     \includegraphics[width=\linewidth]{merged_image_left.png}%
%   }%
%   \hfill
%   \subcaptionbox*{Second subfigure}[.45\linewidth]{%
%     \includegraphics[width=\linewidth]{merged_image_right.png}%
%   }
%   \caption{Two images}
% \end{figure*}

\subsection{Annotation Channels}
\label{subsec:annotation_channels}

Generative models can typically learn and generate high-quality images using the geometric and color information available in images with the traditional red, green, and blue (RGB) channels. 
However, generating high-quality hand images with correct finger count, natural finger positions, and proper finger length poses a unique challenge due to the specific characteristics of hand images discussed in Section~\ref{sec:intro}. To overcome these challenges and to enable generative models to capture the intricate characteristics of hand images, we provide additional information during model training. Consequently, we introduce three annotation channels to provide supplementary information beyond the RGB channels.

\begin{figure*}[h]
    \centering
    \includegraphics[width=1.0\textwidth]{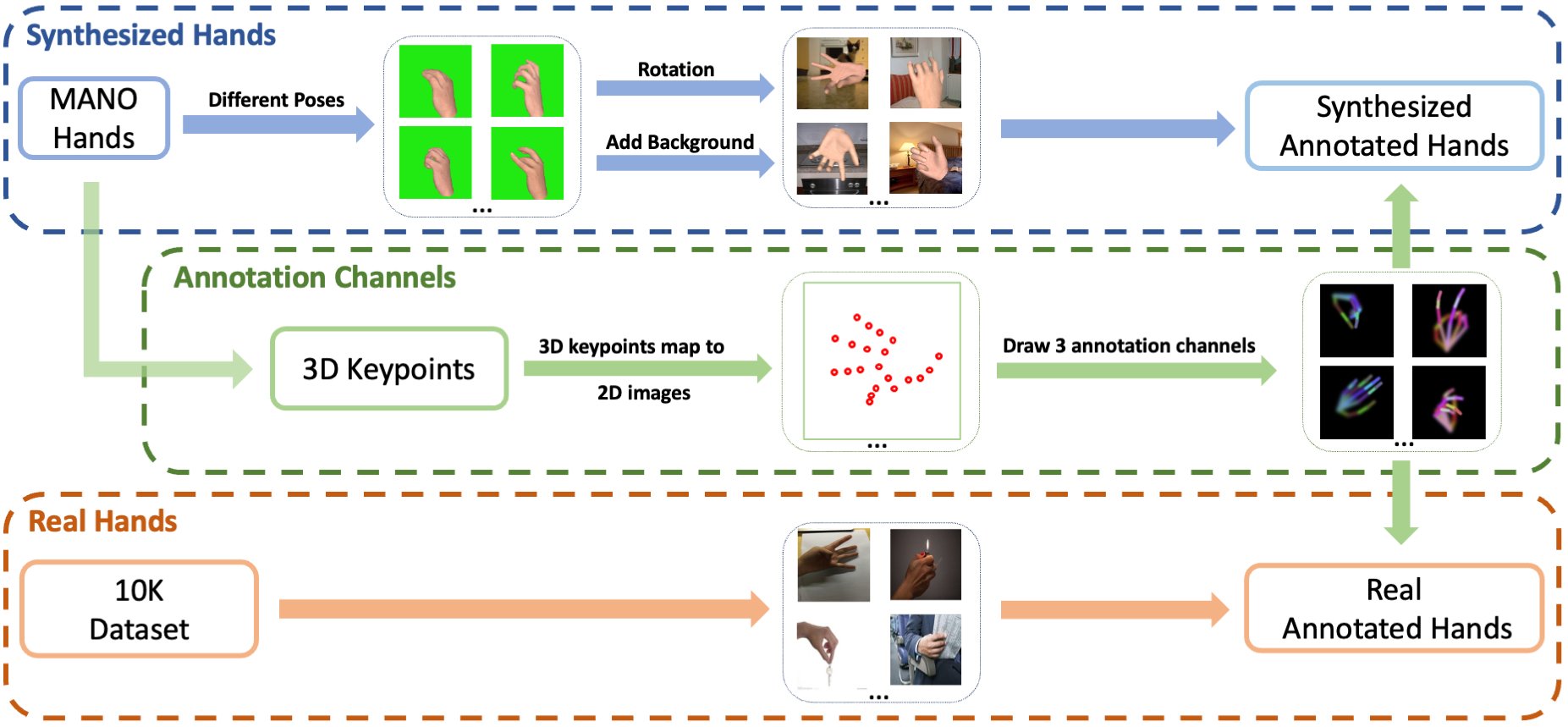}
    \caption{Pipeline for generating annotated hands. To create synthetic hand images, we pass hand size and pose information to the MANO hand model. We then render an image from the model and compose it in front of a background. For real hands, we utilize the Onehand10k dataset and rely on Mediapipe to generate annotations.}
    \label{fig:pipeline}
\end{figure*}

Our annotation channel scheme is centered around producing colored hand skeletons. These skeletons consist of 20 line segments that connect the 21 keypoints of the hand (Figure~\ref{fig:hand_skeleton}), allowing generative models to more easily learn the structure of a hand. By incorporating hand skeletons, we provide valuable geometric information by assigning different colors to different skeleton parts, which imparts additional useful information to the model. This design ensures that the annotation channels deliver essential information to the generative models without significantly modifying the learning process.

In our annotation channel design, we employ three channels that share the same skeleton as a hand but that are indicated with different pixel colors. In our description below, we will assume that the red, green and blue channels are described by integers in the range of 0 to 255.

In the first annotation channel, the grayscale pixel values from the thumb to the little finger take on the values [50, 150, 250, 100, 200], where the segments creating the thumb have pixel values 50 and the segments of the little finger have pixel values 200. (Figure~\ref{fig:1_channel}). This differential pixel value assignment for each finger aids the generative models in distinguishing between individual fingers. Additionally, this annotation layer helps the models learn when to halt the generation of fingers, preventing incorrect finger count in the generated images.

In the second annotation channel, the grayscale pixel values from the hand base to the fingertips are [100, 200, 50, 250]; where the segments connecting to the hand base have pixel value 100 and the fingertip segments have pixel values 250. (Figure~\ref{fig:2_channel}). This pixel value assignment serves to guide generative models in learning the gradual changes from the hand base to the fingertips.

The third annotation layer is dedicated to distinguishing between left and right hands. If the hand in the image is left, we assign all the joints 100  pixel value, while for the right hand we assign 200 pixel value. This layer is crucial as left-right information plays a significant role in generating accurate hand details. As can be seen in Figure~\ref{fig:hand_annotation}, the same skeleton pose could represent the ventral side for the left hand but the dorsal side for the right hand. Substantial differences exist between dorsal and ventral details, such as palmar flexion creases, nails, hand hair, and blood vessels. By incorporating this annotation channel, generative models can more effectively categorize hand details, avoiding the mixing of ventral and dorsal hand features.

\subsection{Synthesized Hands}
\label{subsec:synthesized_hands}

Prior to training on real hand photos, we initially assess the effectiveness of our proposed annotation channels using synthetic data.

% blue box
% todo
As shown in the blue box of Figure~\ref{fig:pipeline}, we leverage the MANO model~\cite{romero2022embodied} to generate synthesized hand images in which the shape, texture and pose of the hand can be varied. The synthesized hand images are rendered with textures~\cite{qian2020html} that encompass variations in age, gender, and ethnicity, thereby ensuring the generation of realistic and visually appealing synthetic hands. In addition to the wide range of finger poses and hand textures, we take additional steps to enhance the variability of our synthesized hand images. This includes comprehensive rotation of the hand in all angles and the integration of diverse backgrounds~\cite{xiao2010sun} representing various scenarios like the kitchen, bedroom, bathroom, basement, and more.

To generate annotation channels for synthesized hands, it is necessary to specify 21 hand keypoints along with their corresponding 20 lines to construct a complete hand skeleton. The MANO model provides precise 3-dimensional keypoints information, with which we can determine the corresponding positions of these keypoints within a 2-dimensional image. Once we have obtained the 21 hand keypoints, we use the methods described in Section~\ref{subsec:annotation_channels} to draw skeletons by connecting them. The generation of the first and second annotation channels is straightforward. For the third annotation channel, we can acquire left and right hand information during the synthesis of hands using the MANO model. To ensure proper occlusion, we carefully manage the order in which the skeleton segments are drawn. We draw the farther skeleton segments first, followed by the closer skeleton segments, ensuring that the closer segments properly occlude the farther ones. For identifying the z-dimension of the skeletal joint, the mean of the z-dimension of the connecting keypoints are taken and then we draw the skeletal joints in the order of farthest to nearest. 

We have created a synthetic hands dataset with 10,000 image samples using the aforementioned method. This data was used to train our models on synthetic hand images. We plan to make this dataset publicaly available.

% \textcolor{red}{@Atith, please add 1-2 sentences here to describe the details.}

\subsection{Real Hands}
\label{subsec:real_hands}

% Mediapipe keypoints detector -> 2D keypoints
% Mediapipe left and right hand classifier -> left or right

The large-scale generative models are typically trained on real photographs from the internet. These models are expected to generate high quality images of people along with real looking hands; however, even the popular and commercial generative models fail to generate consistently good looking hands. 

For training with real hand images we use the Onehand10k dataset consisting of 10,000 images \cite{8529221}. The dataset consists of hand images in real life scenarios, with a wide range of variations in terms of rotation, gesture, poses, and representation in terms of age, gender, and ethnicity. 

For generating annotations of the real hands we use mediapipe, a powerful and popular library for hands identification \cite{lugaresi2019mediapipe}. When hands are processed using mediapipe, it identifies landmarks for each of the 21 keypoints of hands in 3D. Once we have the 3D positions of all the keypoints, we use the method described in Section~\ref{subsec:annotation_channels} to draw the segments by joining the specific keypoints of the hands. Using the keypoint orderings obtained by mediapipe, the position of the joint segments can be identified and colored accordingly for generating the first and the second annotation channels. For third channel, we get the handedness (left or right) information from mediapipe. 

In some of the images, the hands were partially present, and so the joint predicted by mediapipe were outside the image grid. We discarded those images for which all the hand joints were not present inside the image. Our final training dataset consisted of 9931 images.

% \textcolor{red}{@Atith, please add 1-2 paragraphs here to describe the content in orange box of Fig 5. Also, talk about how you form the 3rd annotation channel (i.e., left and right hand).}
\section{Experiments and Results}

% \textcolor{red}{todo - overview}
We evaluate the proposed method on both Generative Adversarial Networks and the diffusion model in Section~\ref{subsec:gan_diffusion}.

% Include something about this later in the results section:

% todo - place it later on

% Real hand dataset being limited and the image backgrounds having high variability, the trained diffusion model generated poor images images for both 3 channels and the 6 channels dataset. We have decided to not include the real hand results in the main paper; however, we have discussed about it in the appendix.

To measure the quality of generated hand images, we design three new metrics as described in Section~\ref{subsec:metric_design}. Finally, we show synthesized hands results and comparative quantitative results in Section~\ref{subsec:results}. 

We evaluate the performance of the GAN model on both real and synthetic hand images, and for diffusion model we test our approach only on synthetic hand images. We evaluate the performances of the model on 2000 images generated using seed 1-2000.

\subsection{Design of GAN Models and Diffusion Models}
\label{subsec:gan_diffusion}

To assess the effectiveness of our proposed method, we employ two types of generative models, namely GAN and the Diffusion Model, known for their exceptional capabilities in image generation, to assess the effectiveness of our proposed methods.

The architecture of the GAN model we employ is derived from StyleGAN2~\cite{karras2020analyzing}, recognized for its impressive image generation performance. In adapting the StyleGAN2 model to our methodology, we made a key modification by expanding the input channel from 3 to 6. This straightforward architectural adjustment holds broad applicability beyond StyleGAN2 and can be readily extended to other GAN models, thus underscoring the versatility of our approach in catering to various downstream models. Below are the training details for StyleGAN2:

% \begin{table}
% \centering
%   \resizebox{0.3\textwidth}{!}{
%   \begin{tabular}{cc}
%     \hline
%     \textbf{Property} & \textbf{Value}\\
    
%     \hline
    
%     Image Size & 256 x 256\\
%     gamma value & 50.0\\
%     Epochs & 2048 \\
%     Batch size & 4 \\
    
%     \begin{tabular}{@{}c@{}}Generator \\ (Reg Interval)\end{tabular}
%      & 4 \\
%       \begin{tabular}{@{}l@{}}Discriminator \\ (Reg Interval)\end{tabular}
%      & 16 \\
%      GPU & RTX\_6000 \\
%      Num of GPUs & 1 \\

%   \hline
% \end{tabular}
% }
%   \caption{Details of the model architecture.}
%   \label{tab:architecture}

% \end{table}

The diffusion model architecture we employ is based on~\citep{denoising-diffusion-pytorch}. The embedded U-Net architecture follows an encoder-decoder structure with a base dimension of 64 and multiplicative factors of (1, 2, 4, 8).

\subsection{Design of Hand Evaluation Metrics}
\label{subsec:metric_design}
% Method-1: Classifier (Overall)
% Method-2: Keypoint detector confidence (key nodes)
% Method-3: FID & Inception Score

The Fréchet Inception Distance (FID)~\cite{heusel2017gans} is a widely used method for evaluating the quality of images created by generative models. However, considering the unique features of hands, it becomes necessary for us to develop new approaches tailored to particularly assess the quality of generated hand images. In addition to FID, we introduce several new metrics that are specifically devised for hands.

\subsubsection{Mediapipe Confidence}

Mediapipe detects 21 keypoints of hands in an image. It also assigns a  hand detection confidence score with each hand predicted in the image. If the hands are not clear, blurry or there are no hands in the image, Mediapipe will still assign 21 keypoints but with less confidence. If the hands are recognizable in the image, Mediapipe assigns a score with a higher confidence. Thus, Mediapipe hand keypoints detection confidence is one of the key metrics that we use to test the efficacy of our approach. 

\subsubsection{Above 90\% Confidence}

In addition to hand detection confidence, we assessed the percentage of generated hands deemed of high quality, using a 90\% confidence threshold. The ``Above 90\% Confidence'' metric reports the percentage of generated hands with a Mediapipe hand detection confidence greater than or equal to 90\%.

% Apart from getting point detection confidence, we also evaluate the proportion of generated images having above 90\% hand detector confidence \textcolor{red}{[maybe we should add more description of how the hand detector confidence is generated?]}. \textcolor{red}{[I don't quite get this sentence]} This is an important metric because even images with good hand detector confidence may not look realistic. 

%With the recent boom in generative models, there are many open-source and subscription based generative models available in the market. Whenever these models doesn't generate hands properly the users rely on in-painiting/editing the generated image where they keep on asking the model to regenerate the area with defective hands, till good quality hands are not generated. Thus, identifying how much proportion of hands are of good quality is an important metric to evaluate these models on the hand generation task.

\subsubsection{Mean Joint Ratio Difference}

Even though existing generative models can create images of hands that are recognizable, the finger segment lengths may not be in a proper ratio with each other. For instance, it is common for humans to have the middle finger longer than the other fingers, and the pinky shorter than all other fingers. While there may be occasional exceptions, the lengths of the average human fingers generally adhere to a specific ratio. To achieve lifelike hands, we assess our method through mean joint ratio difference. This involves obtaining the mean joint length of generated hands, normalizing the joint vector, and calculating the mean squared difference with the normalized mean joint length of hands in the dataset.

Let $H_i$ be the joint lengths for the $i$-th generated hand, where $H_i = [h_{i,1}, h_{i,2}, \ldots, h_{i,20}]$ is a list of 20 joints.

Let $\overline{H}$ be the mean of hand joint lengths for all generated hands, where $\overline{H} = [\overline{h}_1, \overline{h}_2, \ldots, \overline{h}_{20}]$.

Let $N_g$ be the normalized mean joint length vector for the generated hands, where $N_g = \left[\frac{\overline{h}_1}{\|\overline{H}\|}, \frac{\overline{h}_2}{\|\overline{H}\|}, \ldots, \frac{\overline{h}_{20}}{\|\overline{H}\|}\right]$.

Similarly, we calculate $N_d$ be the normalized mean joint vector for the hands in the dataset.

Then,

\[
\text{Mean Joint Ratio Difference} = \sqrt{\sum_{i=1}^{21} (N_d[i] - N_g[i])^2}
\]

\subsection{Results}
\label{subsec:results}

\subsubsection{StyleGAN2}
\label{sec:stylegan2results}
\begin{table}
\centering
  \resizebox{0.5\textwidth}{!}{
  \begin{tabular}{lllll}
    \hline
     Dataset & FID  & Mediapipe  & Above 90$\%$ & Mean Joints\\
            & Score $\downarrow$ & Confidence $\uparrow$ & Confidence $\uparrow$ & Ratio Differences $\downarrow$ \\
     % & Channels & Real & Synthetic  \\
    
    \hline
    
    \begin{tabular}{@{}c@{}}Synthetic \\ (3 channels)\end{tabular}  & \textbf{55.54} & 0.573 & 0.361 & 0.0476 \\
    % \hline
    \begin{tabular}{@{}c@{}}Synthetic \\ (6 channels)\end{tabular} & 81.94 & \textbf{0.729} & \textbf{0.367} & \textbf{0.0178}\\
    % \hline

    \begin{tabular}{@{}c@{}}Real \\ (3 channels)\end{tabular} & 164.75 & 0.4328 & 0.25 & 0.0367 \\
    % (3 channels) &  &  & &  \\
    % \hline
    \begin{tabular}{@{}c@{}}Real \\ (6 channels)\end{tabular} & \textbf{133.63} & \textbf{0.693} & \textbf{0.443} & \textbf{0.0293} \\
    \hline

\end{tabular}
}

\vspace{10pt}

  \caption{StyleGAN2 results trained over 3 and 6 channels, both for synthetic (top) and real (bottom) training images.  }
  \label{tab:StyleGAN2_accuracy}
\end{table}

Table~\ref{tab:StyleGAN2_accuracy} lists StyleGAN2 results from training on both 6-channel and 3-channel datasets, and assesses these results using the different evaluation metrics. The \textit{Mediapipe Confidence} is notably higher for the 6-channel model in both real and synthetic hands when compared to the 3-channel model. The introduction of annotations during training increases the percentage of images with Mediapipe hand keypoint detection confidence, achieving \textit{Above 90\% Confidence}. The \textit{Mean Joint Ratio Difference} metric measures the average deviation of joint length ratios, calculated via Mediapipe keypoints, in generated hands from those in the dataset. Notably, the model trained with 6-channel data produces images with a lower (i.e., better) \textit{Mean Joint Ratio Difference} than its 3-channel counterpart, signifying a closer match in joint ratios to the ground truth dataset hands.

For the real hands dataset, the 6-channel model outperforms the 3-channel model in terms of the \textit{FID score}. However, in the synthetic hands dataset, the 3-channel model achieves a superior \textit{FID score}. The synthetic dataset includes hand images with backgrounds, and we suspect that our 6-channel model prioritizes generating hands, as the additional channels exclusively focus on hand features. The \textit{FID score}, which measures the distribution difference between generated and dataset images, is influenced by the backgrounds generated by the model. Therefore, this dynamic results in a lower (i.e., better) \textit{FID score} for the 3-channel model in the synthetic dataset, as it captures both hand and background information.

\begin{figure*}[h]
    \centering
    \includegraphics[width=1.0\textwidth]{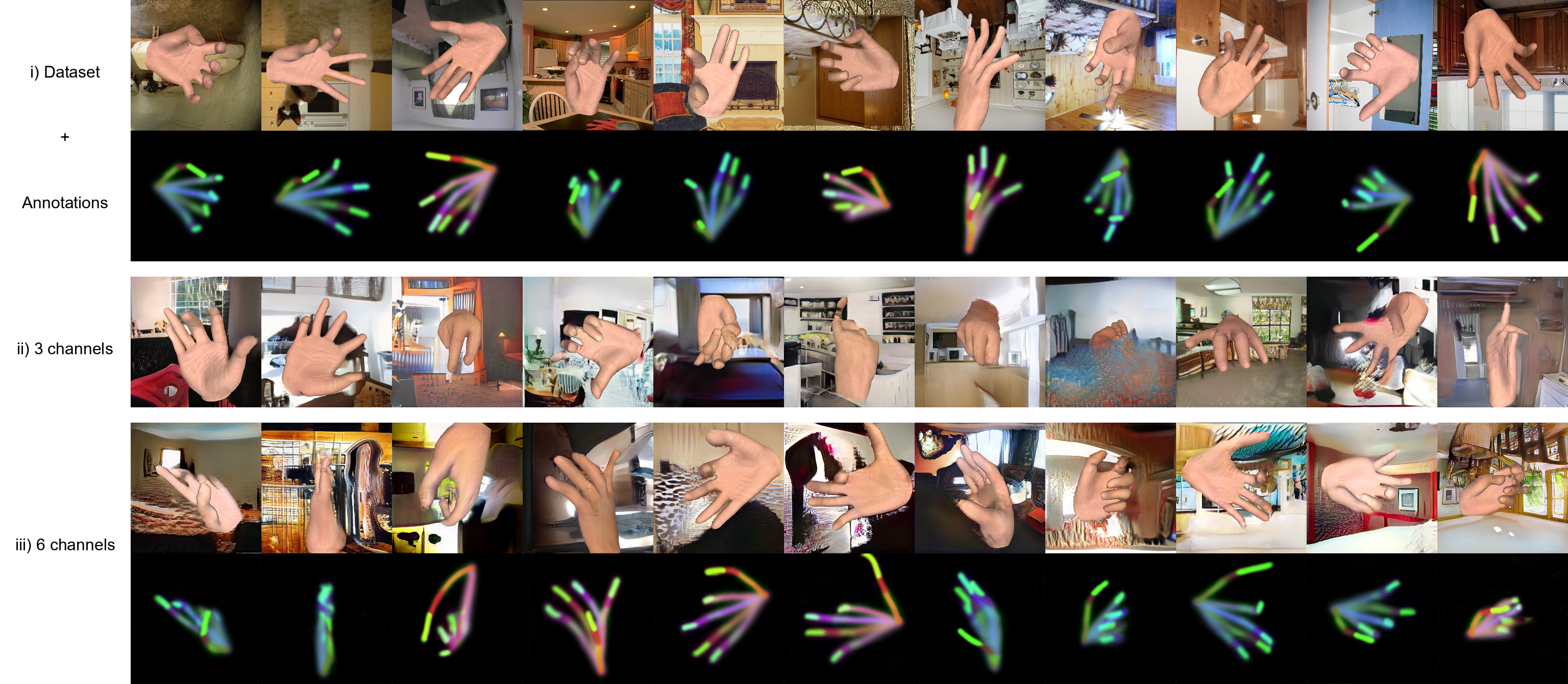}
    \caption{Synthetic Hand images generated by StyleGAN2 when trained over 6 channels and 3 channels (1) The images from the dataset + corresponding annotations, (2) The images generated after training the StyleGAN2 model over synthetic hand images with background (3) The images and the corrseponding annotations generated after training the StyleGAN2 model over 6 channel synthetic hand images with background}
    \label{fig:Stylegan2_synthetic_images}
\end{figure*}

\begin{figure*}[h]
    \centering
    \includegraphics[width=1.0\textwidth]{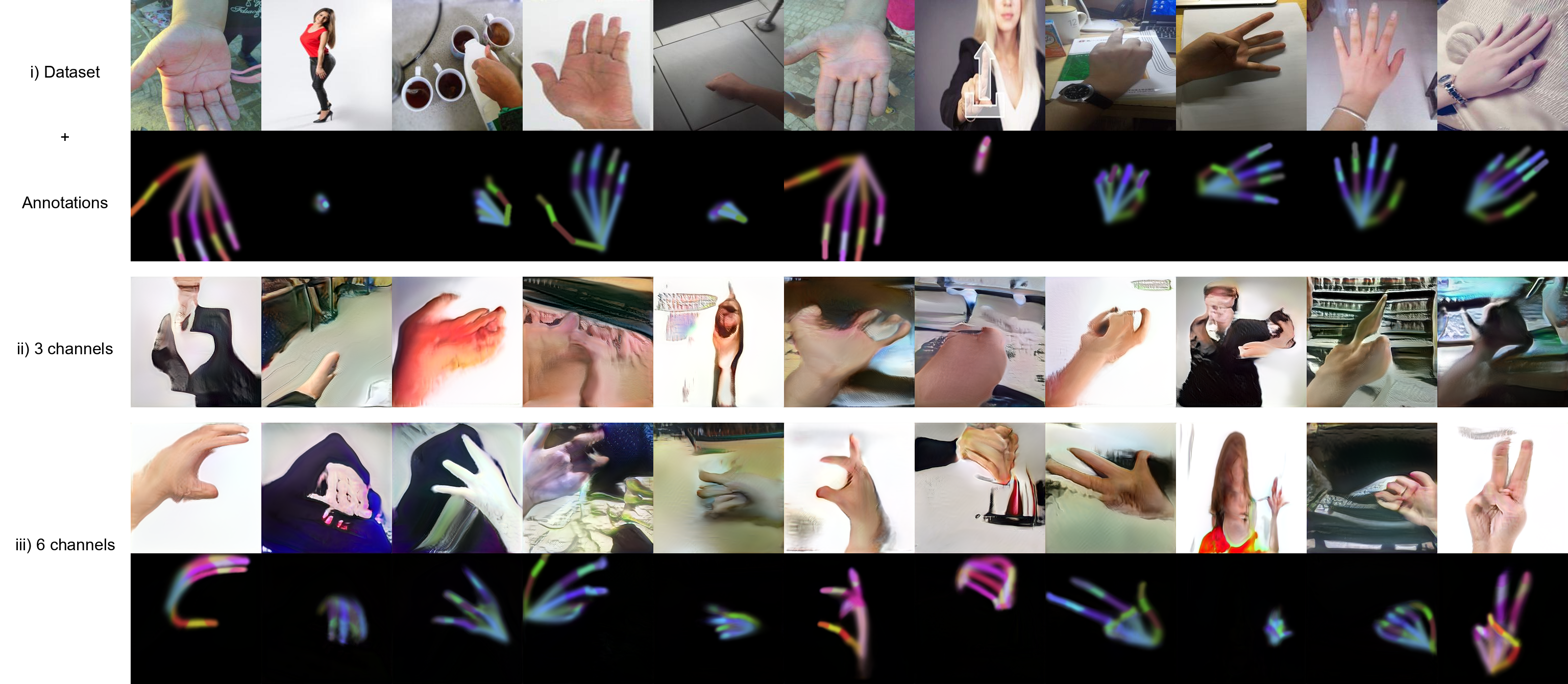}
    \caption{Real Hand images generated by StyleGAN2 when trained over 6 channels and 3 channels (1) The images from the real hands dataset + corrseponding annotations (2) The images generated after training the StyleGAN2 model over real hand images (3) The images and the corrseponding annotations generated after training the StyleGAN2 model over 6 channeled real hand images}
    \label{fig:Stylegan2_real_images}
\end{figure*}

To qualitatively compare the hands generated by StyleGAN2 trained with 3-channel annotated data, please consult Figure \ref{fig:Stylegan2_synthetic_images} for synthetic hands and Figure \ref{fig:Stylegan2_real_images} for real hands.

% Table \ref{tab:StyleGAN2_top90}, compares the probability of getting above 90\% Mediapipe hand detection confidence. The StyleGAN2 model is able to generate higher proportion of images with above 90\% accuracy when trained over 6 channels.

% Finally, we compare the difference in the hand joints ratio vector with the hands in the dataset in table \ref{tab:StyleGAN2_joints}. In this metric we only consider the hands for which we have got above 90\% keypoint detection confidence. For real hands the joint ratio is more aligned to the hands in the dataset for 6 channels than 3 channels. However, for synthetic hands the hand joint ratio is more aligned to the dataset for 3 channels. Similarly the hand joints are more aligned for 6 channels in synthetic hand dataset.
\subsubsection{Diffusion Model}
\begin{table}
\centering
  \resizebox{0.5\textwidth}{!}{

  \begin{tabular}{lllll}
    \hline
     Dataset & FID  & Mediapipe  & Above 90$\%$ & Mean Joints\\
            & Score $\downarrow$ & Confidence $\uparrow$ & Confidence $\uparrow$ & Ratio Difference $\downarrow$ \\
     % & Channels & Real & Synthetic  \\
    
    % \hline

    % \begin{tabular}{@{}c@{}}Real \\ (3 channels)\end{tabular} &  \textbf{154.815} & 0.170 & 0.0103 & \textbf{0.0364}  \\
    % \hline
    % \begin{tabular}{@{}c@{}}Real \\ (6 channels)\end{tabular} & 194.052 & \textbf{0.352} & \textbf{0.0202} & 0.0436\\
    % \hline
    \hline 
    \begin{tabular}{@{}c@{}}Synthetic \\ (3 channels)\end{tabular} & \textbf{134.798} & 0.696 & 0.552 & \textbf{0.015} \\
    % \hline
    \begin{tabular}{@{}c@{}}Synthetic \\ (6 channels)\end{tabular} & 146.781 & \textbf{0.732} & \textbf{0.604} & 0.028\\
    \hline

\end{tabular}

}

    \vspace{10pt}

  \caption{Diffusion model results trained over 3 and 6 channels.  }
  \label{tab:diffusion_accuracy}
\end{table}

% old caption text
% \textit{(i) Mediapipe Accuracy} - Average Mediapipe Point detection accuracy by Diffusion model when trained over 6 channels and 3 channels for Synthetic and Real Hands dataset \textit{ (ii) Above 90\% confidence} - Percentage of generated images with above 90\% Mediapipe point detection confidence. \textit{(iii) Mean Joint Ratio Difference} - Mean Joint Ratio Difference between the joint length ratios of hands generated by Diffusion model trained using 3 channels and 6 channels plus the hand images in the dataset

Due to constraints on the size of the real hand dataset, coupled with high background variability in its images and limited computing resources, the trained diffusion model produces poor results for both the 3 channels and the 6 channels datasets. Therefore, we are only reporting on synthetic hand results, and exclude the real hand results.

Table \ref{tab:diffusion_accuracy} shows the results over various hand evaluation metrics for the diffusion model trained on the synthetic hand dataset with 3 and 6 channels. The \textit{Mediapipe Confidence} is higher for the model trained with 6 channels compared to 3 channels. Training with 6 channels also increases the percentage of images with Mediapipe hand keypoint detection confidence, achieving \textit{Above 90\% Confidence}.

% In diffusion models also the mean joint ratios of the generated hands using 6 channels are more closer to the mean joint ratio in the dataset.

% \textcolor{black}{[Yue: ``I feel the following statement is inconsistent with Table 2? We should also give some explanation as the FID one?''] In diffusion models also the mean joint ratios of the generated hands using 6 channels are more closer to the mean joint ratio in the dataset.} 

The \textit{FID score} is lower (i.e., better) for the model trained with 3 channels compared to the 6 channels trained model. As detailed in Section~\ref{sec:stylegan2results}, the synthetic hand dataset comprises hand images with backgrounds. The 6-channel model prioritizes generating hands over background images due to the additional channels dedicated to hand structure. This prioritization contributes to the lower (i.e., better) \textit{FID score} for the 3 channels model.

The \textit{Mean Joint Ratio Difference} is lower for 3 channels than for 6 channels, indicating that the diffusion model generates a higher proportion of hands that are less recognizable compared to StyleGAN2. Mediapipe, having been trained on a large set of hands, possesses prior knowledge of hand joints and their ratios. In instances where hands are not distinctly recognizable, we suspect that Mediapipe, uncertain about the presence of hands, projects keypoints based on its understanding of the given image and knowledge of hand structures. Consequently, in scenarios where generated hands lack clarity, Mediapipe tends to produce keypoints with joint segments featuring better ratios, albeit with lower confidence.

\begin{figure*}[h]
    \centering
    \includegraphics[width=1.0\textwidth]{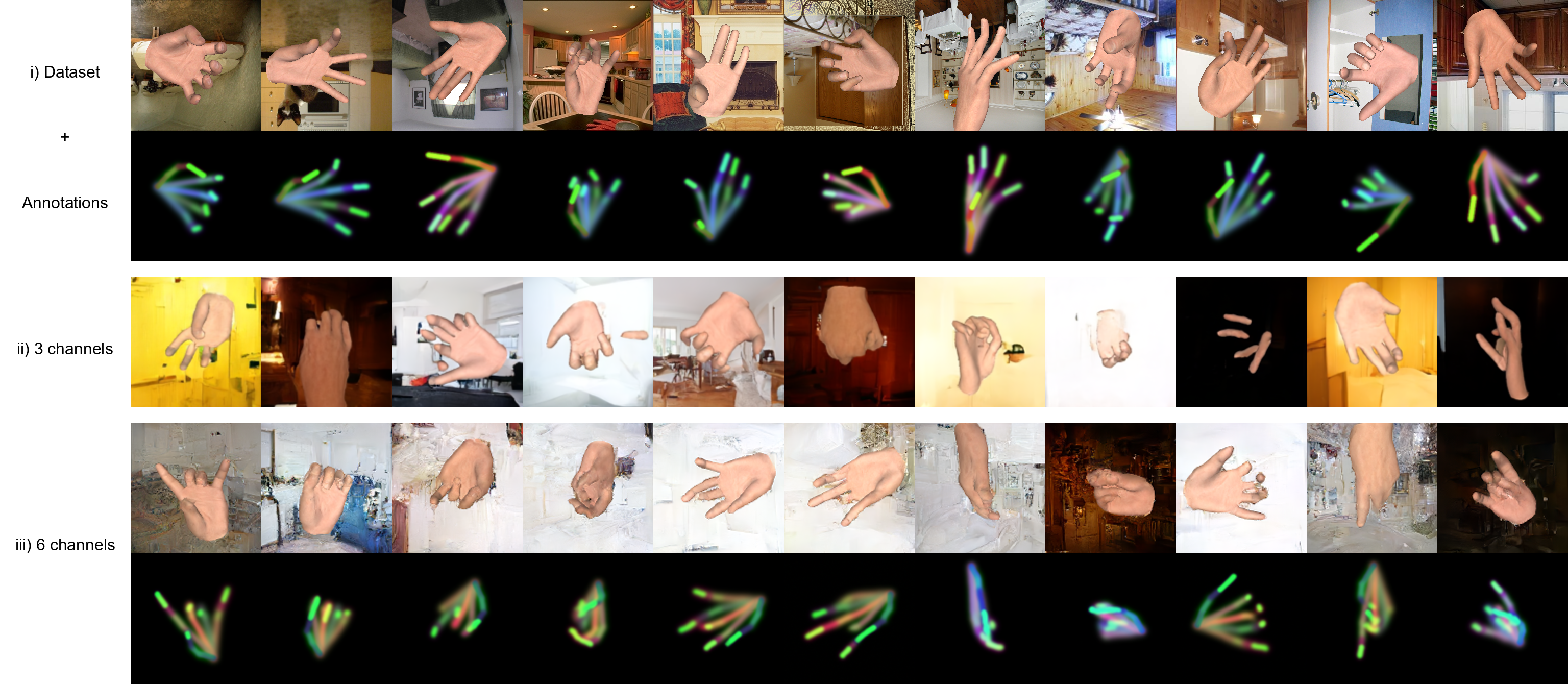}
    \caption{Synthetic hand images generated by diffusion model when trained over 6 channels and 3 channels. (1) The images from the synthetic hands dataset + corrseponding annotations (2) The images generated after training the Diffusion model over synthetic hand images with background (3) The images and the corrseponding annotations generated after training the Diffusion model over 6 channeled synthetic hand images with background}
    \label{fig:Diffusion_synthetic_images}
\end{figure*}

For see the results of synthetic hands generated by the Diffusion model trained with 3 channels and annotated data, please refer to Figure \ref{fig:Diffusion_synthetic_images}.

\begin{table}
\centering
  \resizebox{0.45\textwidth}{!}{
  \begin{tabular}{lcc}
    \hline
   
     No. of Channels & Mediapipe Confidence \\
    
    \hline
    
    3 channels & 0.433  \\
    3 channels + 4th channel & 0.606  \\
    3 channels + 5th channel & 0.620  \\
    3 channels + 6th channel & 0.601   \\
    6 channels & 0.693 \\
     % Exponential & 0.884 \small{\textpm 0.006)} & 0.849 \small{(\textpm 0.007)} \\
     % EWC & 0.743 \small{(\textpm 0.024)} & 0.523 \small{(\textpm 0.016)} \\
     % EWC (online) & 0.670 \small{(\textpm 0.016)} & 0.444 \small{(\textpm 0.013)}\\
     % Lower baseline & 0.472 \small{(\textpm 0.007)} & 0.424 \small{(\textpm 0.012)}\\
    \hline

\end{tabular}
}

    \vspace{10pt}

  \caption{Average \textit{Mediapipe Accuracy} in StyleGAN2 was evaluated across various training configurations on real hands dataset: 3 channels, (3 + 4th) channels, (3 + 5th) channels, (3 + 6th) channels, and all 6 channels.}
  \label{tab:StyleGAN2_ablation_studies}
\end{table}

\subsubsection{Ablation Studies}

We conducted ablation studies to explore how measured accuracy changes with the individual use of different annotation channels in addition to the original image (RGB channel). Table \ref{tab:StyleGAN2_ablation_studies} presents StyleGAN2 results when trained using real hand datasets with RGB plus one of the 4th, 5th, or 6th channels on real hands dataset. The findings indicate that the inclusion of any annotation channel enhances \textit{Mediapipe Confidence}. The accuracy over the 3-channel dataset is approximately 0.4326, and with the addition of just one annotation channel, it improves to around 0.600. Notably, training with all 6 channels yields higher accuracy than training with only 3 plus any one additional channel, suggesting that utilizing all 3 channels results in a more significant performance improvement. Additionally, the 5th channel proves to be the most effective, as it demonstrates a more substantial increase in accuracy compared to the other channels. For the 4th and 6th channels, a similar increase in accuracy is observed when either of the channels is individually added along with the original image.
\section{Discussion and Limitations}

Our approach to the problem of hand generation is to alter the training of the generative model. Our method does not require any special user input during image generation. There are other methods that require user intervention to create images that contain improved hands. For example, some diffusion models have been trained to allow \emph{inpainting}, where the user marks a region of an image and the generative model then fills in the marked regions in a manner that matches its surroundings. If the model generates an image in which a hand looks unrealistic, the user can invoke inpainting to attempt to fix the hand. This method not only requires user judgement, but sometimes several inpainting attempts are required before a satisfactory result is generated.

Another approach to address the issue of poor hand generation for a diffusion model is to use ControlNet~\cite{zhang2023adding}. ControlNet requires a user to provide a guiding image, and then uses features detected from such an image as part of the control that is used to generate another similar image. Several kinds of features can be used to guide image generation, including detected edges, skeleton segments or depth. To generate images with better hands, the user would need to provide an image that has a hand in the same pose as they desire in a generated image. While this method can produce good resulting hands, this requires the user to provide features from a control image for every new image that is to be generated. This requires considerably more intervention on the user's part than simply providing a text prompt. We note that our annotated channels look quite similar to the skeleton features from the OpenPose module in ControlNet. It is important to recognize that we use such color-labeled skeleton features only during model training, whereas ControlNet requires an OpenPose skeleton to be provided for each image that is to be generated when this module is being used.

The biggest limitation to our work is the lack of a larger database of images that contain hands. The onehand10k database that we used is modest in size compared to the large databases that are used to train diffusion models. For example, Stable Diffusion 1.5 was trained on the 600 million image dataset called LAION Aesthetics 5+~\cite{sd1.5, laion_aesthetics}. It would be ideal if we could use our method to train on such a large database of images that includes hands in a wide variety of contexts. Some of these contexts might include hand and faces, hands as part of whole body images of people, and hands interacting with a variety of objects. Unfortunately, gathering such a database would be an extensive undertaking. Creating the labels for such a varied collection of images might also require automatic hand identification that is more robust under these broader contexts.

Another limitation of our work is that we are unable to train large diffusion models due to lack of compute resources. While there are several publicly available diffusion models that have already been trained (e.g. Stable Diffusion), the more capable of these models required months of training on supercomputer clusters. Our own modest academic compute resources do not allow us to train a diffusion model at such scale. As indicated in Table~\ref{table:hyperparam-diffusion} in the Appendix, the training of the diffusion model requires approximately 10 days (on one Nvidia RTX 4090 Ti), even for a modest 5 epochs.

Finally, the mediapipe library has its own limitations. We are relying on mediapipe from annotation generations in real hands and evaluations. Although, overall mediapipe does good work in hand keypoint detections and handedness (left or right) identification, still it is not 100\% accurate. Our evaluation metrics like "Mediapipe Confidence", "Above 90\% Confidence", and "Mean Joint Ratio Difference" are based on mediapipe. Changing the underlying library for our evaluation metrics may impact the outcomes. 
\section{Conclusion}

We have demonstrated that by training a generative model with additional annotated image channels, we can improve the quality of hand image generation. Our method is not model specific, and can be used to improve hand image generation using any generative model paradigm (VAEs, GANs, diffusion models, and so on).

There are several possible directions for future work. First, we would like to carry out larger scale training using many more annotated hand images. We have not yet undertaken the task of collecting a larger hand image dataset. Second, it would be ideal to train a large text-to-image diffusion model using our method for annotating hands. Finally, our general approach to annotation may be applicable to other 3D objects that have high degree of freedom configurations.

% Bibliography
\bibliographystyle{ACM-Reference-Format}
\bibliography{refs}

\clearpage

% Appendix
\appendix
\section{Training Details}

The training hyper-parameters for StyleGAN are included in Table~\ref{table:stylegan}. The training hyper-parameters for diffusion model are included in Table~\ref{table:hyperparam-diffusion}.

\begin{table}[h]
\centering
\begin{tabular}{|c|c|}
\hline
\textbf{Parameter} & \textbf{Value} \\ \hline
learning rate                 & 1e-4                           \\ \hline
batch size                    &     64                      \\ \hline
image size & $256 \times 256$ \\  \hline
training sample count - Real Hands & 9,931 \\  \hline
training clock time & $\sim$ 120 hr \\  \hline
epochs & 2048\\ \hline
gamma value & 50 \\ \hline

\end{tabular}
\caption{Hyper-parameters for StyleGAN2 model.}
\label{table:stylegan}
\end{table}

% todo - 1/3 epoch (say diffusion model takes much more time - check the time it takes) & image size

\begin{table}[h]
\centering
\begin{tabular}{|c|c|}
\hline
\textbf{Parameter} & \textbf{Value} \\ \hline
learning rate                 & 8e-5                           \\ \hline
batch size                    & 32                             \\ \hline
image size & $128 \times 128$ \\  \hline
training sample count & 10,000 \\  \hline
training clock time & $\sim$240 hr \\  \hline
epochs & $\sim$5 \\ \hline
timesteps per image              & 1000                           \\ \hline
total timesteps               & 1,500,000                        \\ \hline
\end{tabular}
\caption{Hyper-parameters for diffusion model.}
\label{table:hyperparam-diffusion}
\end{table}

\newpage

\section{More Results}

We include more results on subsequent pages from our annotation approach, both from StyleGAN2 and from a simple diffusion model.

\subsection{StyleGAN2 - Synthetic Hands}

\begin{figure*}[h]
    \centering
    \includegraphics[width=0.9\textwidth]{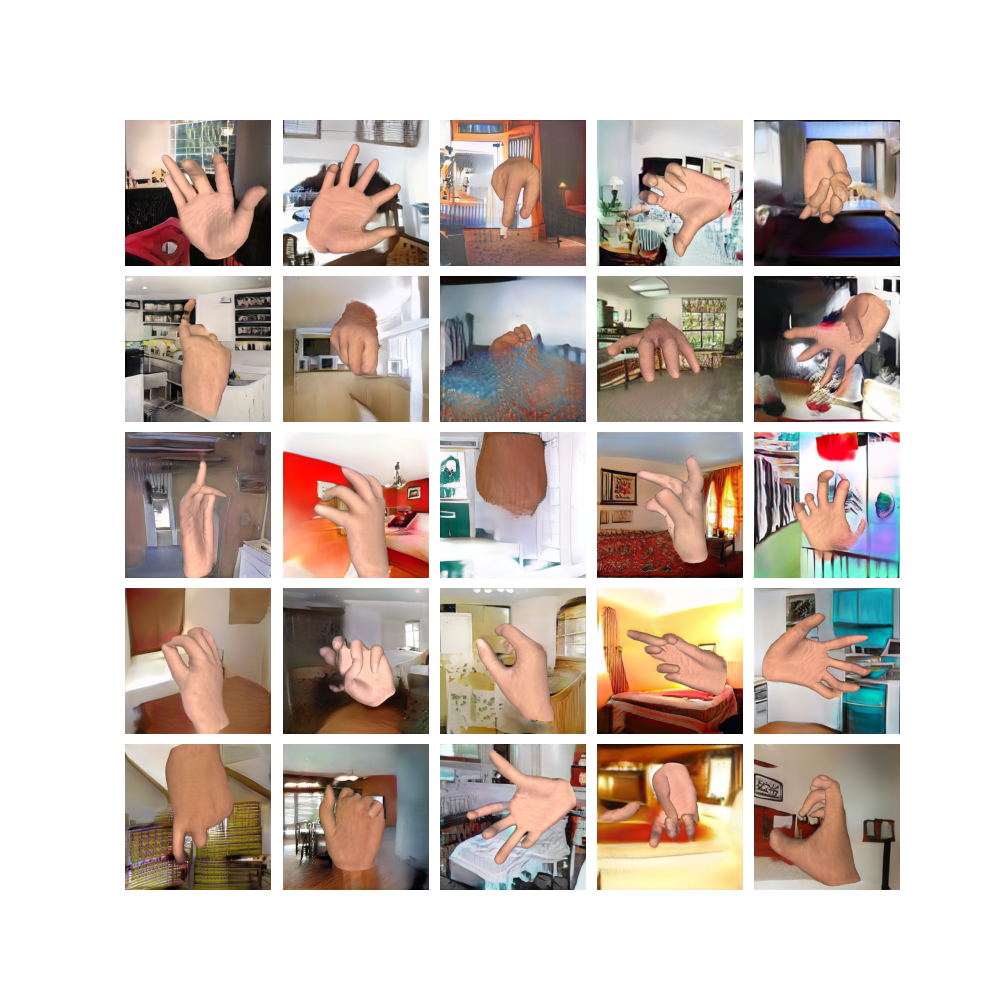}
    \caption{Generated images from SyleGAN2 that was trained on synthetic hand data with 3 channels. Seeds used 100 - 124}
    \label{fig:appendix_stylegan2_synthetic_3_channels_images}
\end{figure*}

\begin{figure*}[h]
    \centering
    \includegraphics[width=0.9\textwidth]{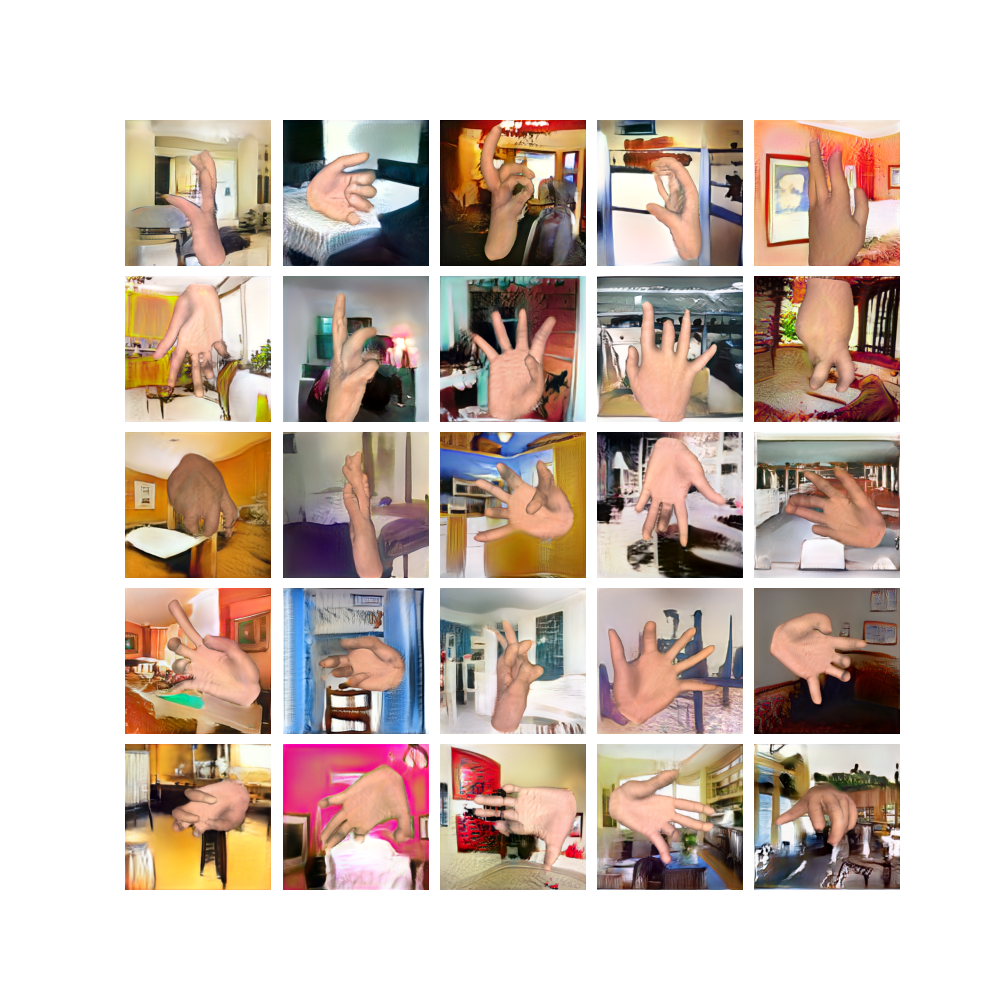}
    \caption{Generated images from StyleGAN2 model that was trained on synthetic hand data with 6 channels. Seeds used 100 - 124}
    \label{fig:appendix_stylegan2_synthetic_6_channels_images}
\end{figure*}

Refer - Figure \ref{fig:appendix_stylegan2_synthetic_3_channels_images} for images generated using 3 channels dataset, and Figure \ref{fig:appendix_stylegan2_synthetic_6_channels_images} for images generated using the annotated dataset.

\subsection{StyleGAN2 - Real Hands}
\begin{figure*}[h]
    \centering
    \includegraphics[width=0.9\textwidth]{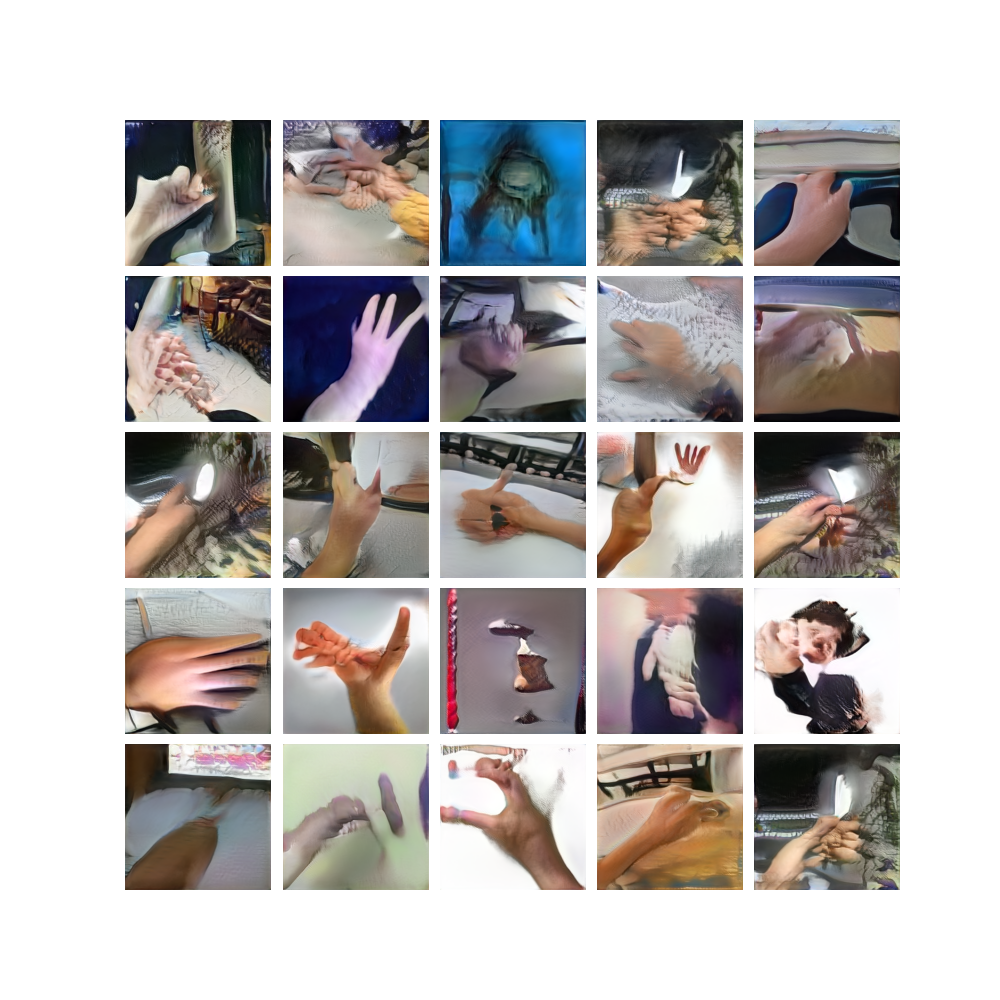}
    \caption{Generated images from StyleGAN2 model that was trained on real hand data with 3 channels. Seeds used 100 - 124}
    \label{fig:appendix_stylegan2_real_3_channels_images}
\end{figure*}

\begin{figure*}[h]
    \centering
    \includegraphics[width=0.9\textwidth]{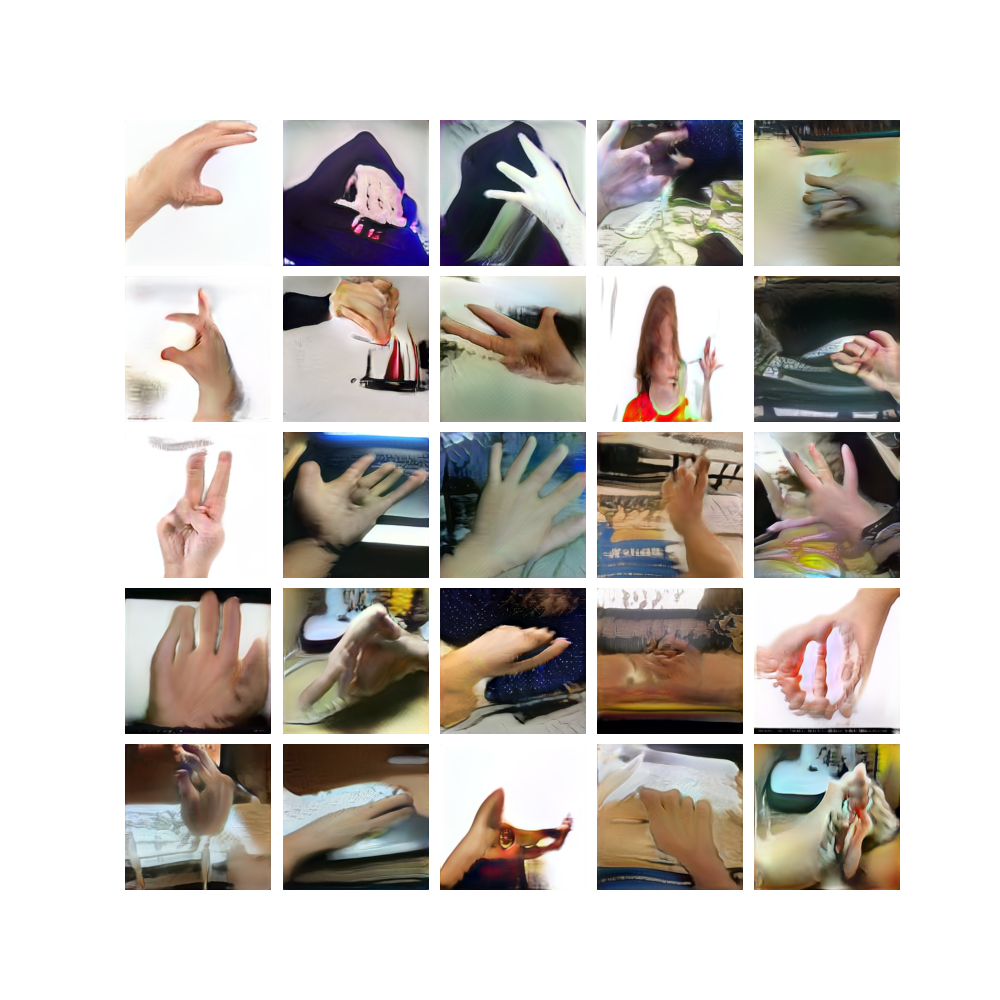}
    \caption{Generated images from SyleGAN2 that was trained on real hand data with 6 channels. Seeds used 100 - 124}
    \label{fig:appendix_stylegan2_real_6_channels_images}
\end{figure*}

Refer - Figure \ref{fig:appendix_stylegan2_real_3_channels_images} for images generated using 3 channels dataset, and Figure \ref{fig:appendix_stylegan2_real_6_channels_images} for images generated using the annotated dataset.

% \begin{figure*}[h]
%     \centering
%     \includegraphics[width=0.9\textwidth]{stylegan2_real_grid.png}
%     \caption{Real hands generated by StyleGAN2 trained using 6 channels dataset.}
%     \label{fig:pipeline}
% \end{figure*}
\subsection{Diffusion - Synthetic Hands}

\begin{figure*}[h]
    \centering
    \includegraphics[width=0.9\textwidth]{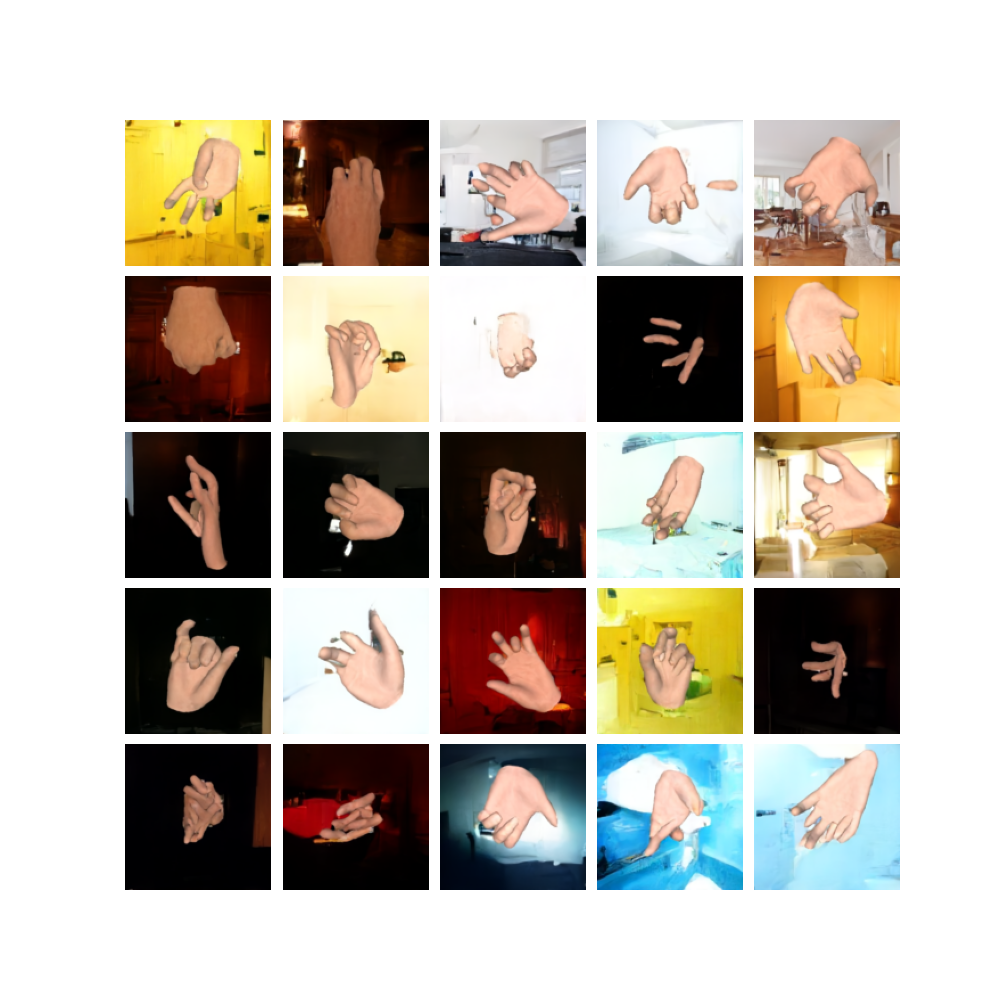}
    \caption{Generated images from a simple diffusion model that was trained on synthetic hand data with 3 channels. Seeds used 100 - 124}
    \label{fig:appendix_diffusion_synthetic_3_channels_images}
\end{figure*}

\begin{figure*}[h]
    \centering
    \includegraphics[width=0.9\textwidth]{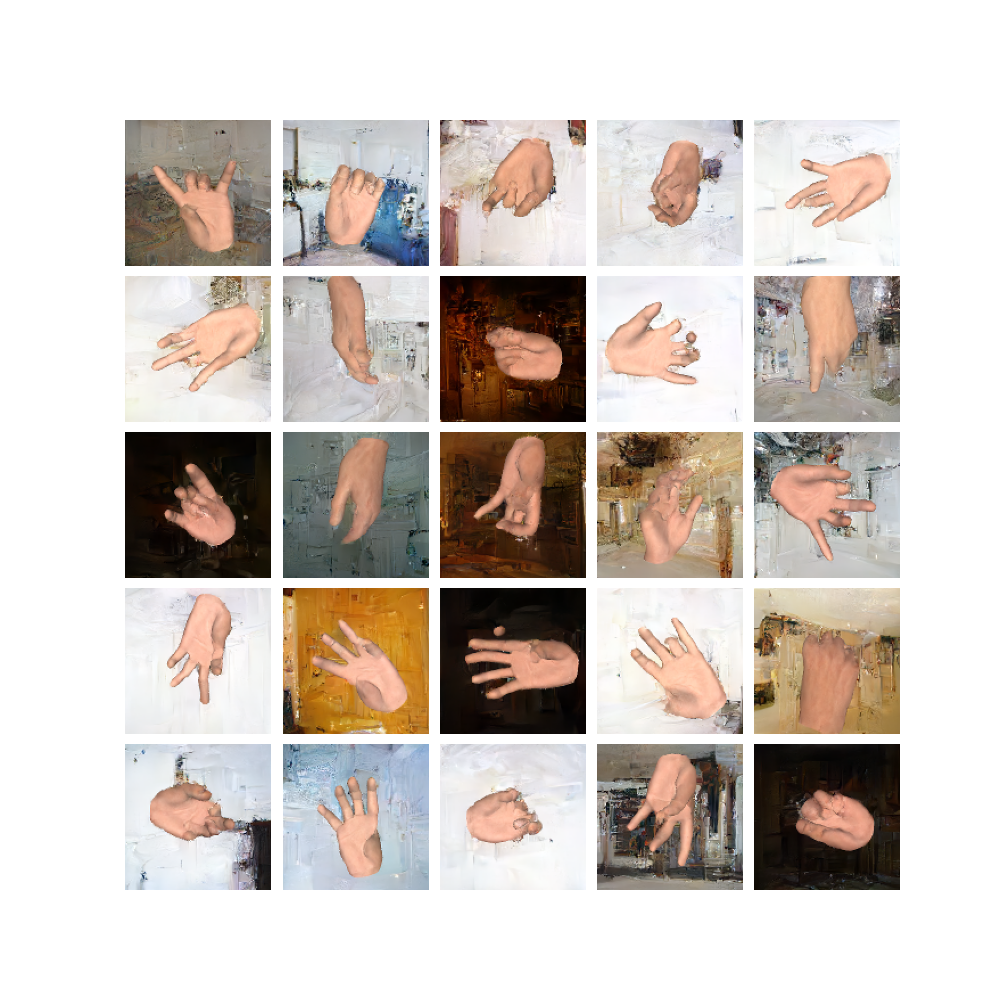}
    \caption{Generated images from a simple diffusion model that was trained on synthetic hand data with 6 channels. Seeds used 100 - 124}
    \label{fig:appendix_diffusion_synthetic_6_channels_images}
\end{figure*}

Refer - Figure \ref{fig:appendix_diffusion_synthetic_3_channels_images} for images generated using 3 channels dataset, and Figure \ref{fig:appendix_diffusion_synthetic_6_channels_images} for images generated using the annotated dataset.

\end{document}